\documentclass[letterpaper, 10 pt, conference]{ieeeconf}  

\IEEEoverridecommandlockouts                              

\usepackage{amsmath} 
\usepackage{amssymb}  

\usepackage{hyperref}

\usepackage{graphicx}
\usepackage{xcolor}
\usepackage{multirow}  
\usepackage{algorithm}
\usepackage[noend]{algpseudocode}

\usepackage{makecell}

\usepackage{hyperref}

\newcommand{\textsfsmall}[1]{\textsf{\small #1}}
\newcommand{\textttsmall}[1]{\texttt{\small #1}}

\title{\LARGE \bf
Sequential Manipulation of Deformable Linear Object Networks with Endpoint Pose Measurements using Adaptive Model Predictive Control
}

\author{Tyler Toner$^{1,3, *}$, Vahidreza Molazadeh$^{3}$, Miguel Saez$^{3}$, Dawn M. Tilbury$^{1,2}$, and Kira Barton$^{1,2}$  
\thanks{* Tyler Toner is the corresponding author: \href{mailto:twtoner@umich.edu}{twtoner@umich.edu}}
\thanks{$^{1}$Department of Mechanical Engineering, University of Michigan}%
\thanks{$^{2}$Robotics Department, University of Michigan}%
\thanks{$^{3}$General Motors, Research and Development}%
\thanks{See videos and more details at \href{https://sites.google.com/view/robo-harness}{sites.google.com/view/robo-harness}.}
}

\begin{document}

\renewcommand{\arraystretch}{1.5}

\maketitle
\thispagestyle{empty}
\pagestyle{empty}

\begin{abstract}
Robotic manipulation of deformable linear objects (DLOs) is an active area of research, though emerging applications, like automotive wire harness installation, introduce constraints that have not been considered in prior work. Confined workspaces and limited visibility complicate prior assumptions of multi-robot manipulation and direct measurement of DLO configuration (state). This work focuses on single-arm manipulation of stiff DLOs (StDLOs) connected to form a DLO network (DLON), for which the measurements (output) are the endpoint poses of the DLON, which are subject to unknown dynamics during manipulation. To demonstrate feasibility of output-based control without state estimation, direct input-output dynamics are shown to exist by training neural network models on simulated trajectories. Output dynamics are then approximated with polynomials and found to contain well-known rigid body dynamics terms. A composite model consisting of a rigid body model and an online data-driven residual is developed, which predicts output dynamics more accurately than either model alone, and without prior experience with the system. An adaptive model predictive controller is developed with the composite model for DLON manipulation, which completes DLON installation tasks, both in simulation and with a physical automotive wire harness. 
\end{abstract}
\section{Introduction}

Manipulation of deformable objects remains an important challenge in robotics. Many objects of interest fall into the category of deformable \textit{linear} objects (DLOs), including ropes \cite{saha_manipulation_2007}, surgical sutures, \cite{moll_path_2006}, and cables \cite{Jiang2010}. Their deformability is challenging to accurately model, complicating standard manipulation techniques developed for rigid objects. 

Much work has studied the problem of driving a DLO to a particular shape profile or topology \cite{lv_dynamic_2022, wang_offline-online_2022, wu2020learning, yan2021learning, han_model-based_2017, alvarez_interactive_2016, saha_manipulation_2007, moll_path_2006}, sometimes with an emphasis on physical interaction with the environment, such as for knotting \cite{saha_manipulation_2007} or cable routing \cite{fresnillo_approach_2022, zhu_robotic_2020}. In contrast, few works have focused on precisely driving a particular point on a DLO to a goal pose \cite{yu2021adaptive, lim2022_casting, zhou_practical_2020}.
Additionally, most DLO manipulation work has involved multiple cooperating robot arms. This assumption appears in early geometric planning approaches \cite{moll_path_2006, saha_manipulation_2007}, later work focused on practical industrial implementation \cite{Jiang2010, koo2008, fresnillo_approach_2022}, and recent approaches focused on learning and control \cite{wang_offline-online_2022, lv_dynamic_2022, zhou_practical_2020, yu2021adaptive}. Few works have studied single-arm DLO manipulation \cite{han_model-based_2017, alvarez_interactive_2016, lim2022_casting, lv_dynamic_2022}.

Consider the automotive wire harness: a bundle of several heterogeneous, relatively stiff cables, each responsible for delivering power or transmitting data inside a vehicle. Each cable may be considered an individual DLO with unique mechanical properties, terminating with a specific terminal type. The objective of harness installation is to insert each terminal into its respective receptacle in its environment. Due to its complexity, harness manipulation remains a manual task in industry. This task involves some additional constraints compared to those typically considered in DLO literature. 

\begin{figure}
  \centering
  \includegraphics[width=3.4in]{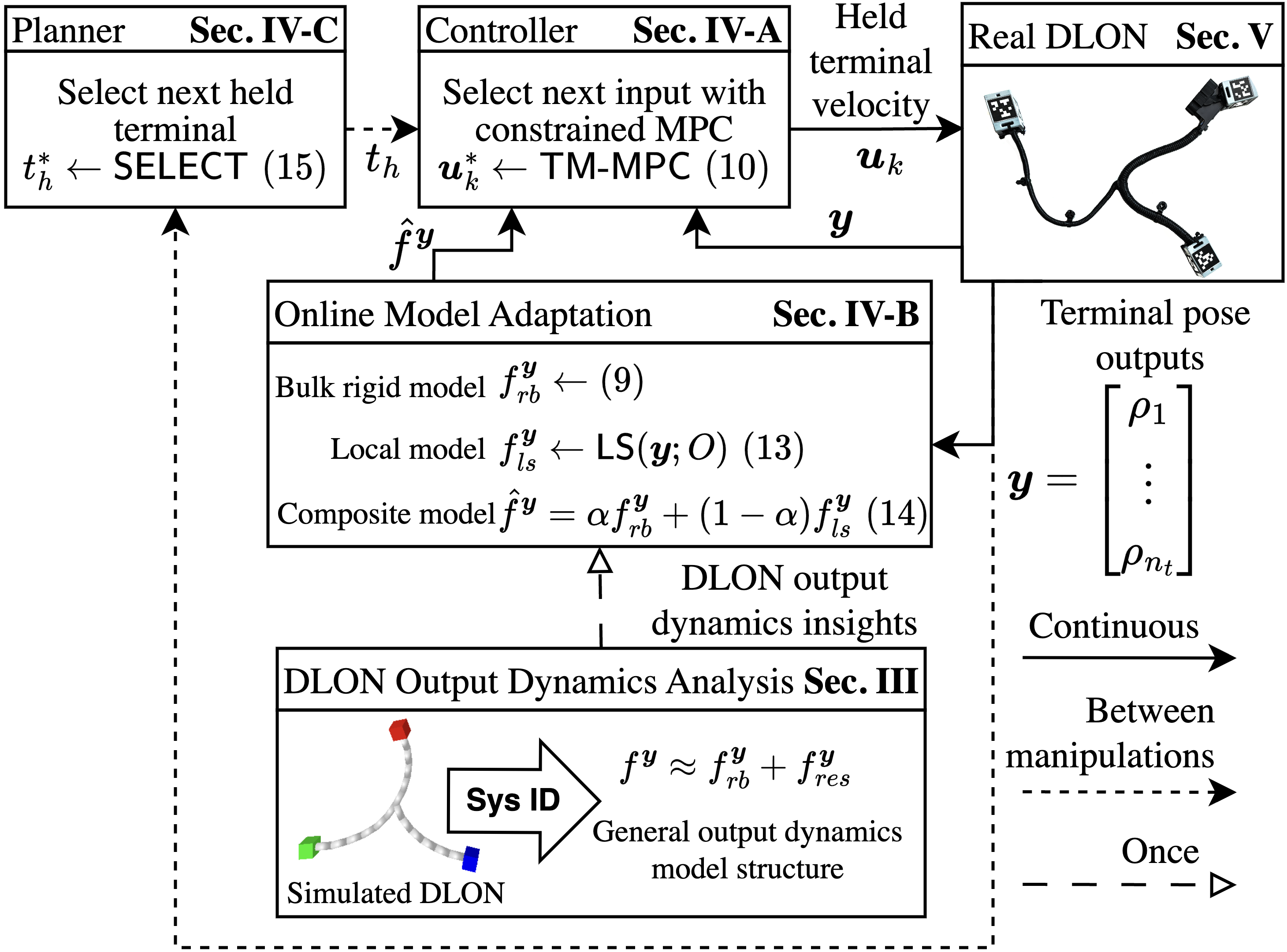}
  \caption{Framework for robotic installation of deformable linear object networks (DLONs), e.g., automotive wire harnesses, using only terminal pose measurements, or outputs. The controller leverages a composite input-output model containing a DLON-agnostic rigid body model and a DLON-specific local model learned \textit{in situ} without prior learning or online excitation.}
  \label{fig: framework overview}
\vspace{-0.7cm}
\end{figure}

First, the wire harness system contains a variety of DLOs, but is not itself a DLO. We term this class of objects \textit{DLO networks} (DLONs). Each DLO in a DLON has a connected end and a free end, or \textit{terminal}. Furthermore, each DLO in the DLON is what we denote as a \textit{stiff DLO} (StDLO), meaning that, for fixed terminal poses, the DLOs themselves do not significantly deform, instead roughly returning to an equilibrium configuration, contrasting DLOs such as strings or threads. Additionally, we specifically consider DLOs with predefined rigid grasp points, termed semi-DLOs (SDLOs) in \cite{zhou_practical_2020}, constraining grasps to terminals. Next, the fundamental objective is to manipulate the DLON in such a way as to bring each terminal to a precise pose goal. This contrasts with the majority of prior work concerned with shape or topology goals. Moreover, the shape profile of a DLON may not be measurable. The limited lighting and reflective surfaces encountered during automotive assembly complicate profile detection using contrast-based segmentation \cite{zhu_dual-arm_2018} or pointcloud processing \cite{wang_offline-online_2022, Nguyen2021}. Instead, limited observation of only key parts of the DLON, like its terminals, may be available. For example, marker-based pose tracking may be used \cite{Jiang2010, koo2008, Toner2023_RL}. Finally, in many real-world scenarios, coordinated multi-arm manipulation is not practical, being typically realized by carefully constructed programs in static, well-modeled environments \cite{Saez2020}. In addition to communication latency and more restrictive self-collision constraints, multi-arm manipulation inside a confined environment like an automotive cabin introduces spatial challenges. 

Thus, the problem considered in this work is single-arm manipulation of planar DLONs with terminal pose goals and measurements. A \textit{manipulation} is a robot motion during which the DLON is grasped, and which ends when the grasp is released. A manipulation of the DLON that brings a terminal to its goal is a \textit{terminal manipulation} (TM). Successful installation requires a sequence of TMs that drive the DLON to a state at which all terminals are simultaneously at their goal poses. We do not consider insertion \cite{zhou_practical_2020, Zhao2022}, but instead assume a reaction force representing insertion is activated upon a successful TM for simplified analysis. 

The work most similar to ours is \cite{lim2022_casting}, which studies planar, single-arm DLO manipulation with endpoint measurements and goals. By leveraging physical experiments to generate simulators, open-loop policies are learned to position the free end of a specific DLO by moving the held end. In contrast, we consider sequential manipulation of multi-terminal DLONs with continuous endpoint pose measurements, which we solve with model predictive control (MPC) using an online adaptive model, allowing enforcement of hard constraints while requiring no offline learning. 

Our objective is a planning and control framework that can solve feasible DLON installation problems, is robust to initial DLON configurations, and operates on novel DLONs without prior training or system identification. The main contributions of this work are:
\begin{enumerate}
    \item Analysis of planar DLON output dynamics, which admit decomposition into simple rigid body dynamics and residual dynamics,

    \item Development of a TM planning and control methodology for TM sequence selection and real-time control using an adaptive composite model that enforces feasibility of future TMs, and
    
    \item Experimental validation of our proposed approach with both a simulator and a physical wire harness. 
\end{enumerate}

Our framework is summarized in Fig. \ref{fig: framework overview}. Output dynamics are studied in section \ref{sec: output dynamics}, revealing a decomposition into a rigid body model $\mathrm{f}^{\boldsymbol{y}}_{rb}$ and residual $\mathrm{f}^{\boldsymbol{y}}_{res}$. An adaptive MPC is designed (section \ref{subsec: mpc}) for constrained TMs using a composite model (section \ref{subsec: online model adaptation}) which leverages the decomposition found in section \ref{subsec: sindy dynamics}. The TM sequence is planned using a simple, yet effective, heuristic (section \ref{subsec: sequence planner}). The planning and control approach is shown to generalize to installation of novel, real DLONs (section \ref{sec: experiments}).

\section{Problem Statement} \label{sec: problem statement}
A DLON consists of a set of connected stiff DLOs (StDLOs), each terminating with one of $n_t$ terminals. The objective of DLON installation is to bring each terminal to its respective receptacle. During a terminal manipulation (TM), if a \textit{held} terminal is successfully brought to its receptacle, it is \textit{mated} and is held in place by a reaction force. Other terminals are \textit{free}, and are subject to harness dynamics.

Due to its flexibility, the configuration, or \textit{state}, of a DLON is not finite, but may be approximated by a set of $n$ features, $\boldsymbol{x} \in \mathbb{R}^n$, for example a set of coordinates along its profile \cite{wang_offline-online_2022, yu2021adaptive, zhu_dual-arm_2018}. Furthermore, we only consider planar DLONs, similar to \cite{wang_offline-online_2022, zhu_robotic_2020, lim2022_casting, Toner2023_RL}. The pose, $\rho = [x  \, \, y \, \, \theta]^\top \in SE(2)$ of each terminal with respect to the robot base frame is continuously monitored by a vision system, resulting in a measurement, or \textit{output}, $\boldsymbol{y} = [\rho_1^\top \cdots \rho_{n_t}^\top ]^\top \in SE(2)^{n_t}$. 

A manipulator can grasp any terminal within a workspace $W \subset SE(2)$ with a minimum clearance of $r_\epsilon$ to the nearest obstacle $o \in O \subset \mathbb{R}^2$. During manipulation, the held terminal is attached to the gripper, which realizes velocity commands $\boldsymbol{u} = [v_x \, \,  v_y \, \,  \omega_z]^\top \in \mathcal{U}$ with negligible dynamics. The held terminal is directly controlled with $\dot{\rho}_{h} = \boldsymbol{u}$, mated terminals are static, $\dot{\rho}_{m} = 0$, and free terminals are subject to:
\begin{gather}
  \boldsymbol{x}_{k+1} = f(\boldsymbol{x}_k, \boldsymbol{u}_k) \label{eq: NL dynamics} \\
  \boldsymbol{y}_k = h(\boldsymbol{x}_k) \label{eq: output map}
\end{gather}

\noindent with discrete time state dynamics $f$ and output map $h$, both of which are unique to a particular DLON. After each TM, $f$ changes, so neither $f$ nor $h$ are assumed \textit{a priori}. 

The problem considered in this work is as follows. \textbf{Given} continuous terminal pose output measurements $\boldsymbol{y}$ of a planar DLON with $n_t$ terminals with goals $\boldsymbol{g} = [\rho_{1,g}^\top \, \, \cdots \, \, \rho_{n_t,g}^\top]^\top $ manipulated by a robot within a workspace $W$ with $r_\epsilon$-constrained grasping in the presence of obstacles $O$, \textbf{find} a sequence of terminal manipulations that drive all terminals to their goals, $\boldsymbol{y} \rightarrow \boldsymbol{g}$. We make the following assumptions:

\begin{enumerate}
    \item [(A1)] $\exists$ no explicit constraints on $\boldsymbol{x}$, only $\boldsymbol{y}$, and

    \item [(A2)] over any TM, $\| \dot{\boldsymbol{x}} \|$ is sufficiently small that the system can be modeled as quasi-static.
    
\end{enumerate}

\section{Output Dynamics} \label{sec: output dynamics}

To control a DLON subject to constraints, a dynamic model of its free terminals trajectories in response to held terminal velocity inputs $\boldsymbol{u}$ is critical. As both $f$ and $h$ are unknown, and the state $\boldsymbol{x}$ is not available, a prerequisite for output-based control is to determine whether terminal pose output dynamics can be predicted with output $\boldsymbol{y}$ measurements alone; in other words, direct output dynamics $\boldsymbol{y}_{k+1} = f^{\boldsymbol{y}}(\boldsymbol{y}_k, \boldsymbol{u}_k)$ must exist. To answer this question, a simulated DLON is used to gather a dataset of of $\boldsymbol{x}$ and $\boldsymbol{y}$ (section \ref{subsec: dataset collection}), from which we find an approximate inverse to the output map $h$ (section \ref{subsec: output invertibility}), indicating the existence of $f^{\boldsymbol{y}}$. Output dynamics are approximated by a polynomial model (section \ref{subsec: sindy dynamics}), revealing a decomposition later leveraged for an adaptive control-oriented model (section \ref{subsec: online model adaptation}). 

\subsection{DLON simulation and dataset collection} \label{subsec: dataset collection}

\begin{figure}[]
  \centering
  \vspace{0.25cm}
  \includegraphics[width=3.3in]{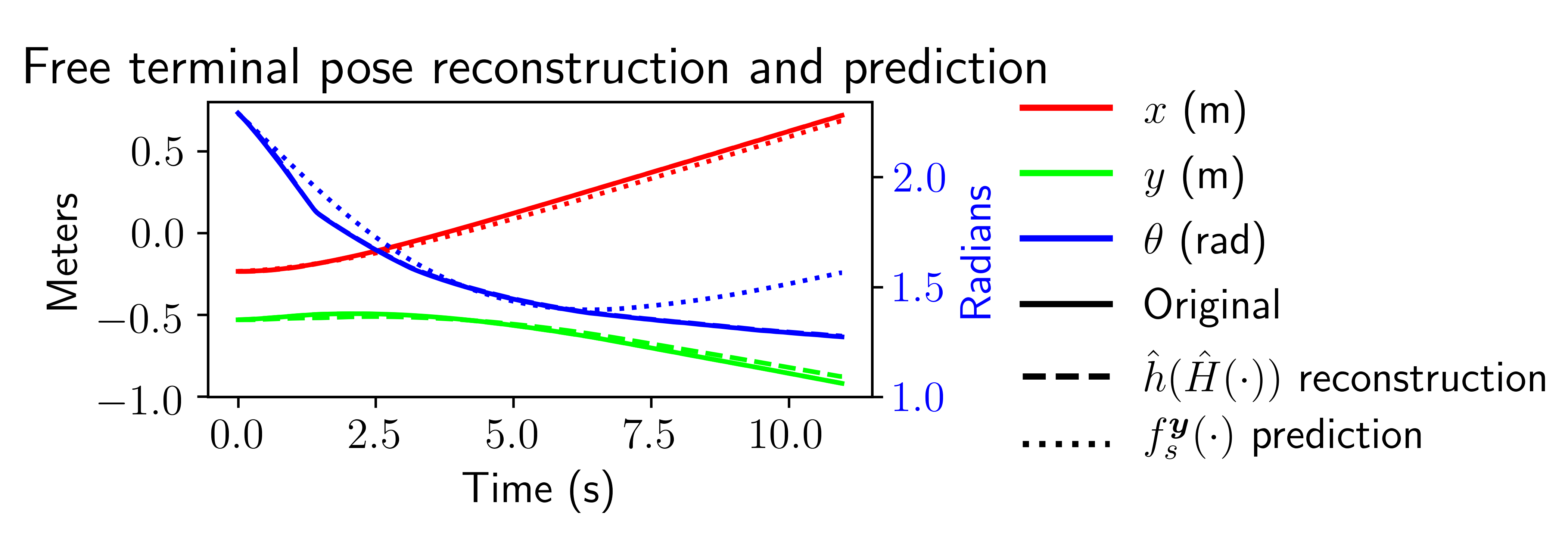}
  \caption{Representative pose trajectory $\rho_{t,k} = [x_{t,k} \, \, y_{t,k} \, \, \theta_{t,k}]^\top$ of one free terminal $t$, selected from $\mathcal{D}$ (solid lines). The reconstruction of $\rho_{t,k}$ using the output map network and its inverse network, $\hat{h}(\hat{H}(\cdot))$ (dashed lines) is close enough to the original to be mostly obscured. The prediction of $\rho_{t,k}$ from $\rho_{t,0}$ using the polynomial model $f^{\boldsymbol{y}}_{s}$ (dotted lines) tracks the original initially, but deviates in $\theta$ after several seconds.}
  \label{fig: all model trajectories}
  \vspace{-0.5cm}
\end{figure}

To study DLON behavior, a simulated proxy (Fig. \ref{fig: simulated experiments}) of a real harness (Fig. \ref{fig: physical experiments}(a)) was developed with the goal of emulating its qualitative behavior. A DLON model was generated with three branches, roughly 650 mm in the longest direction, terminating with $n_t = 3$ cuboid terminals with 45 mm side lengths. Flexibility of the intermediate DLOs was achieved by arranging $n_L$ rigid cylindrical links connected by spherical joints (reducing to $z$ axis revolute joints during $x$-$y$ planar manipulation), each subject to internal damping, constrained angular displacement, and friction. 

The DLON was simulated in Pybullet \cite{coumans2019pybullet}, allowing continuous measurement of joint angles $\theta_j$ link poses $\rho$, as well as application of forces and torques at the $F_s = 240$ Hz simulation rate. Controllers were developed to generate forces and torques that track reference position and velocity commands. In each simulation, terminal $t = 0$ was held and commanded at desired velocities $\boldsymbol{u}$. To explore the state space, the DLON was excited with constant $\boldsymbol{u}$ commands, uniformly sampled from the admissible control space $\mathcal{U}$, for 15 seconds. The DLON was initialized with the same initial conditions, but the beginning of each trajectory was removed in order to ensure the initial state of each trajectory is unique.

Trajectories of $\theta_j$ and $\boldsymbol{y} = [\rho_0^\top \, \, \rho_1^\top \, \, \rho_2^\top]^\top$ were recorded and filtered, with velocities $\dot{\boldsymbol{y}}$ estimated using central differences. Each recording was downsampled to a realistic $F_c = 30$ Hz. With $\boldsymbol{x} = [\rho_0^\top \, \,  \theta_1 \, \,  \dots \, \theta_{n_L} ]^\top$, the uniform-length trajectories were formed into the dataset:
\begin{equation} \label{eqn: dataset}
    \mathcal{D} = \{ \{ \boldsymbol{x}_k, \boldsymbol{y}_k, \dot{\boldsymbol{y}}_{k}, \boldsymbol{u}_k\}_{k=0}^{N_s} \}.
\end{equation}

\subsection{Output invertibility} \label{subsec: output invertibility}

Although DLONs vary in terms of structure, stiffness, and other physical properties, we qualitatively observe from real and simulated systems that wire harnesses have unique stable states, the defining characteristic of a StDLOs. For a fixed set of terminal poses $\overline{\boldsymbol{y}}$, the DLON tends to return to a unique state $\overline{\boldsymbol{x}}$ when external forces along the DLON are removed. Stable, minimum energy configurations of DLOs have been studied in other work and serve as a useful analytical simplification \cite{moll_path_2006}. We hypothesize that there exists
\begin{equation} \label{eqn: approximate output map}
    H : SE(2)^{n_t} \rightarrow \mathcal{B}(\boldsymbol{x}, \epsilon_x) \subset \mathbb{R}^n
\end{equation}

\noindent that approximates the inverse of $h$, such that 
\begin{equation} \label{eqn: approximate inverse}
    \| h(H(\boldsymbol{y})) - \boldsymbol{y} \| < \epsilon_y
\end{equation}

\noindent where $\mathcal{B}(\boldsymbol{x}, \epsilon_x)$ is a ball centered at $\boldsymbol{x}$ with radius $\epsilon_x$, and $\epsilon_x, \epsilon_y > 0$ are thresholds of acceptable precision. To evaluate this hypothesis, we utilize the dataset $\mathcal{D}$ gathered in section \ref{subsec: dataset collection} to learn neural network approximations $\hat{h}$ and $\hat{H}$ using a standard supervised learning approach. 

To aid training, outputs were altered to contain poses of free terminals relative to the held terminal and the rotation re-encoded as $\boldsymbol{y} = [ ({}^0 \rho_1^{sc})^\top \, \,  ({}^0 \rho_2^{sc})^\top ]^\top$ with $\rho^{sc} = [ x \, \, y \, \, \sin(\theta) \, \,  \cos(\theta)]^\top$. A 70\%-15\%-15\% dataset split was used to separate training, validation, and testing data. Multilayer perceptrons were used for both $\hat{h}$ and $\hat{H}$. Optuna \cite{Akiba2019} was used for hyperparameter tuning with minimization of $L_2$ loss on the validation set as the tuning objective. 

Error metrics of the final networks are shown on the test dataset in Table \ref{table: network errors}, which indicate $\hat{h}$ and $\hat{H}$ have approximated both the output map and its inverse successfully. The output reconstruction error $\| \hat{h}(\hat{H}(\boldsymbol{y})) - \boldsymbol{y} \|$ is likewise small, indicating that $\hat{H}$ approximates the output of $\hat{h}$ as desired.  Moreover, applying the two networks to reconstruct a representative trajectory illustrates the fit (Fig. \ref{fig: all model trajectories}). 

Substituting \eqref{eq: NL dynamics} into \eqref{eq: output map}, we have $\boldsymbol{y}_{k+1} = h(f(\boldsymbol{x}_k, \boldsymbol{u}_k))$. Supposing that $H$ exists, it follows that:
\begin{equation}
  \boldsymbol{y}_{k+1} \approx h(f(H(\boldsymbol{y}_k), \boldsymbol{u}_k)) = f^{\boldsymbol{y}}(\boldsymbol{y}_k, \boldsymbol{u}_k),
\end{equation}

\noindent indicating the existence of direct input-output dynamics $f^{\boldsymbol{y}}$, allowing output trajectories to be uniquely predicted without knowledge of the underlying state.  

\subsection{Sparse identification of output dynamics} \label{subsec: sindy dynamics}

\begin{table}[]
    \centering
    \vspace{0.25cm}
    \caption{Mean model prediction errors}
    \begin{tabular}{cccc}
        \hline
        \makecell{Model error metric} & \makecell{Number of \\ parameters} & \makecell{Translational \\ Error (m)} & \makecell{Rotational \\ Error (rad)} \\
        \hline
        $\|\hat{h}(\boldsymbol{x}) - \boldsymbol{y}\| $ & 5631 & $1.50 \times 10^{-3}$ & $1.53 \times 10^{-3}$ \\
        $\|\hat{H}(\boldsymbol{y}) - \boldsymbol{x}\| $ & 9939 & N/A & $5.19 \times 10^{-2}$ \\
        $\| \hat{h}(\hat{H}(\boldsymbol{y})) - \boldsymbol{y}\|$ & N/A & $2.08 \times 10^{-3}$ & $5.28 \times 10^{-3}$ \\
         $\| f^{\boldsymbol{y}}_{s}(\boldsymbol{y}_k) - \boldsymbol{y}_{k+1}\|$ & 110 & $3.57 \times 10^{-4}$ & $1.09 \times 10^{-3}$ \\
        \hline
    \end{tabular}
    \label{table: network errors}
    \vspace{-0.5cm}
\end{table}

Next, we seek an interpretable model of $f^{\boldsymbol{y}}$ capable of providing insight into the structure of DLON output dynamics, and turn to the sparse identification of nonlinear dynamics (SINDy) technique \cite{brunton_discovering_2016_sindy, brunton_sparse_2016_sindyc}. Consider a library of nonlinear function candidates $\boldsymbol{\Theta}(\boldsymbol{y}, \boldsymbol{u})$, in which each column represents a nonlinear function of $\boldsymbol{y}$ and $\boldsymbol{u}$. For a continuous time system of the form $\dot{\boldsymbol{y}} = \mathrm{f}^{\boldsymbol{y}}(\boldsymbol{y}, \boldsymbol{u})$, SINDy seeks an approximation $\mathrm{f}^{\boldsymbol{y}}(\boldsymbol{y}, \boldsymbol{u}) \approx \mathrm{f}^{\boldsymbol{y}}_{s}(\boldsymbol{y}, \boldsymbol{u}) = \boldsymbol{\Theta}(\boldsymbol{y}, \boldsymbol{u}) \mathbf{\Xi}$ containing as few terms as possible. To find $\mathbf{\Xi}^*$, solve:
\begin{equation} \label{eq: sindy}
    \min_{\mathbf{\Xi}} \sum_{\mathcal{D}}\left\| \dot{\boldsymbol{y}}_k - \boldsymbol{\Theta}(\boldsymbol{y}_k, \boldsymbol{u}_k) \mathbf{\Xi} \right\|_2^2 + \lambda \|\mathbf{\Xi}\|_1 \, \, \text{s.t.} \, \, \min |\mathbf{\Xi}| \geq T
\end{equation}

\noindent where $\lambda > 0$ promotes a sparse solution and $T$ is the minimum coefficient threshold to avoid miniscule terms. 

The PySINDy package \cite{de2020pysindy} was used to solve \eqref{eq: sindy}. A simple polynomial library on $\boldsymbol{y}, \boldsymbol{u}$ terms with maximum degree 2 was utilized. Once again, the dataset $\mathcal{D}$ was split and Optuna was utilized to select optimal $\lambda, T$ while solving \eqref{eq: sindy} over the training set, in order to minimize prediction error on the validation set. The resulting continuous time model $\mathrm{f}_{s}^{\boldsymbol{y}}(\boldsymbol{y}, \boldsymbol{u}) = \boldsymbol{\Theta}(\boldsymbol{y}, \boldsymbol{u}) \mathbf{\Xi}^*$ was discretized by assuming a zero-order hold over the time step $\Delta t$: $f^{\boldsymbol{y}}_s(\boldsymbol{y}, \boldsymbol{u}) = \boldsymbol{y} + \Delta t \, \mathrm{f}_{s}^{\boldsymbol{y}}(\boldsymbol{y}, \boldsymbol{u})$. An optimal model with only 110 terms was found using $\lambda = 0.957$, $T = 0.108$.

The polynomial model $f^{\boldsymbol{y}}_s$  was evaluated in terms of one-step prediction accuracy in Table \ref{table: network errors} and trajectory prediction performance over 11 seconds in Fig. \ref{fig: all model trajectories}. $f^{\boldsymbol{y}}_s$ predicts translational dynamics well, with reduced performance in the rotational dimension. Despite the imperfect fit, the polynomial model is interpretable; upon inspection, we observe that it can be decomposed into two distinct components:
\begin{equation} \label{eq:sindy_rb_residual}
   \mathrm{f}^{\boldsymbol{y}}_{s}(\boldsymbol{y}, \boldsymbol{u}) = C \mathrm{f}^{\boldsymbol{y}}_{rb}(\boldsymbol{y}, \boldsymbol{u}) + \mathrm{f}^{\boldsymbol{y}}_{res}(\boldsymbol{y}, \boldsymbol{u})
\end{equation}

\noindent where $\mathrm{f}^{\boldsymbol{y}}_{rb}(\boldsymbol{y}, \boldsymbol{u})$ are rigid body dynamics with diagonal $C \succ 0$ and $\mathrm{f}^{\boldsymbol{y}}_{res}(\boldsymbol{y}, \boldsymbol{u})$ are residual terms. Rigid body dynamics between the held terminal $h$ and a terminal $t$ are given by:
\begin{equation} \label{eq: rigid body dynamics}
  \mathrm{f}^{\rho_t}_{rb}(\dot{\rho}_t, \boldsymbol{u}) = B_{rb}(\rho_t) \boldsymbol{u} = 
  \begin{bmatrix}
      1 & 0 & y_h - y_t \\
      0 & 1 & x_t - x_h \\
      0 & 0 & 1 \\
  \end{bmatrix} 
  \begin{bmatrix}
      v_x \\ v_y \\ \omega_z
  \end{bmatrix}
\end{equation}

\noindent which can be stacked for: $\mathrm{f}^{\boldsymbol{y}}_{rb} = [ \mathrm{f}^{\rho_1 \top}_{rb}  \cdots  \mathrm{f}^{\rho_{n_t} \top}_{rb} ]^\top$. Note that \eqref{eq: rigid body dynamics} is a classical equation for rigid body motion in a plane, and is not specific to any particular DLON. 

To evaluate how well rigid body dynamics $\mathrm{f}^{\boldsymbol{y}}_{rb}$ predict the true continuous-time dynamics $\mathrm{f}^{\boldsymbol{y}}$, both were computed over all samples in the dataset $\mathcal{D}$, with $\mathrm{f}^{\boldsymbol{y}}(\boldsymbol{y}_k, \boldsymbol{u}_k) \approx \dot{\boldsymbol{y}}_k$, and the the coefficient of determination, or $R^2$, was computed between all values of $\mathrm{f}^{\boldsymbol{y}}_{rb}$ and $\dot{\boldsymbol{y}}_k$. Averaging over both free terminals, translational and rotational $R^2$ values were found to be 0.972 and 0.477, respectively, indicating that rigid body dynamics \eqref{eq: rigid body dynamics} are significantly predictive of translational dynamics and moderately predictive of rotational dynamics. 

Although $f$ and $h$ are unknown, output dynamics were found to be well-approximated by a polynomial model containing significant rigid body motion. Bulk output dynamics, available \textit{a priori} for any DLON, may be predicted as rigid, with comparatively minor DLON-specific residuals. These results are based on the a single simulated DLON dataset, but we see the resulting adaptive model derived from these modeling insights (section \ref{subsec: online model adaptation}) generalizes to a physical DLON (section \ref{fig: physical experiments}). Limitations are discussed in section \ref{sec: conclusions}. 


\section{Planning and Control} \label{sec: planning and control}

\begin{table}
    \centering
    \vspace{0.25cm}
    \caption{Maximum trajectory prediction error ($\textrm{mean} \pm \textrm{std}$)}
    \label{table: ensemble model prediction error}
    \tabcolsep=0.10cm
    \begin{tabular}{lcc}
        \hline
        Model & Translational Error (m) & Rotational Error (rad) \\
        \hline
        Rigid Body $B_{rb}$ \eqref{eq: rigid body dynamics} & $0.065 \pm 0.102 $ & $0.278 \pm 0.532$ \\
        Least Squares $B_{ls}$ \eqref{eq: least squares} & $0.043 \pm 0.037$ & $ 0.174 \pm  0.202 $ \\
        \textbf{Composite} $\hat{B}$ \eqref{eq: ensemble model} & $\mathbf{0.036} \pm \mathbf{0.049}$ & $\mathbf{0.156} \pm \mathbf{0.293}$ \\
        \hline
    \end{tabular}
    \vspace{-0.5cm}
\end{table}

To realize DLON installation, a methodology for planning TM sequences and robustly executing them is developed. For TM execution, an adaptive model predictive control (MPC) approach with appropriate constraints is given in section \ref{subsec: mpc}, with online model adaptation covered in section \ref{subsec: online model adaptation}, which leverages insights from section \ref{subsec: sindy dynamics}. Finally, the TM sequence planning algorithm is given in section \ref{subsec: sequence planner}. 

\subsection{Model predictive control} \label{subsec: mpc}

DLON installation is sequential: each TM must be executed such that future TMs remain feasible, so free terminals must not be driven to ungraspable regions, either due to obstacle proximity or by exiting the robot workspace. As found in section \ref{sec: output dynamics}, simple models predict output dynamics well over short horizons, so we propose an adaptive MPC:
\begin{equation} 
\begin{aligned} \label{eq: mpc problem}
    \textsfsmall{TM-MPC}(\boldsymbol{y}_0; \boldsymbol{g}, O): \, \min_{\boldsymbol{u}_0, \ldots, \boldsymbol{u}_{N-1}} &\sum_{j=0}^{N-1} l(\boldsymbol{y}_j, \boldsymbol{u}_j) + l_t(\boldsymbol{y}_N)  \\
    \textrm{s.t.} \quad & \forall j = 0, \ldots, N \\
        & \boldsymbol{y}_{j+1} = \hat{f}^{\boldsymbol{y}}(\boldsymbol{y}_j, \boldsymbol{u}_j)\\
        & c(\boldsymbol{y}_j) \leq 0 \\
        & \boldsymbol{y}_j \in \mathcal{Y}, \, \boldsymbol{u}_j \in \mathcal{U}
\end{aligned}
\end{equation}

\noindent where \eqref{eq: mpc problem} is solved over a horizon of $N$ future steps to optimize a stage cost $l$ and terminal cost $l_t$ while satisfying constraints: the system dynamics, inequality constraints, and set memberships. Dynamics $\hat{f}^{\boldsymbol{y}}$ are continuously updated online with a procedure covered later in section \ref{subsec: online model adaptation}. At each time step of a TM, output $\boldsymbol{y}_k$ is taken as $\boldsymbol{y}_0$ for \textsfsmall{TM-MPC} \eqref{eq: mpc problem} and the first optimal input $\boldsymbol{u}_k = \boldsymbol{u}_0^*$ is executed. We define an $SE(2)$ metric with rotational weight $\beta$ as:
\begin{equation} \label{eq: se2 distance metric}
 \mathbf{d}_{\beta}(\rho_1, \rho_2) = \left\|  \begin{bmatrix} x_1 - x_2 \\ y_1 - y_2 \end{bmatrix} \right\|_2^2 + \beta \left\|  \begin{bmatrix} \cos(\theta_1) - \cos(\theta_2) \\ \sin(\theta_1) - \sin(\theta_2) \end{bmatrix} \right\|_2^2.
\end{equation}

\noindent Then, we define the costs as 
\begin{equation}  \label{eq: l and lf costs}
\begin{aligned}
  l(\boldsymbol{y}_k, \boldsymbol{u}_k) &= \mathbf{d}_{\beta}(\rho_{h, k}, \rho_{h, g}) + \boldsymbol{u}_k^\top Q \boldsymbol{u}_k + \Delta \boldsymbol{u}_k^\top Q_{\Delta} \Delta \boldsymbol{u}_k \\
  l_t(\boldsymbol{y}_N) &= p \, \mathbf{d}_{\beta}(\rho_{h, N}, \rho_{h,g})
\end{aligned}
\end{equation}

\noindent where $Q, Q_\Delta \succ 0$ and $p \geq 0$ define relative weights on control cost, control increment cost, and terminal cost. The goal pose $\rho_{t,g}$ of each terminal is computed from the pose of its respective receptacle, $\rho_r$ and a predefined insertion offset ${}^r \rho_{t,g}$, both of which are static and known. 

Ensuring future TM feasibility can be conservatively achieved by enforcing free TM graspability at all times. Containment in the reachable workspace is encoded by defining $\mathcal{Y} = W^{n_t}$. A graspability-preserivng clearance of at least $r_\epsilon$ is achieved with $c(\boldsymbol{y})$ by stacking $r_{t,o} - \mathbf{d}_{0}(\rho_{t,k}, \rho_o)$ terms for all obstacles $o$ and terminals $t$. The safety radius $r_{t,o} = r_o + r_t + r_\epsilon$ is defined by circumscribing all terminals and obstacles with circles of radii $r_t$ and $r_o$, respectively.

\subsection{Online model adaptation} \label{subsec: online model adaptation}

To estimate $\hat{f}^{\boldsymbol{y}}$, a common approach is to find a local linear model $\dot{\boldsymbol{y}} = \mathrm{f}_{ls}^{\boldsymbol{y}}(\boldsymbol{u}) = B_{ls} \boldsymbol{u}$ using least squares:
\begin{equation}   \label{eq: least squares}
    \textsfsmall{LS}(\dot{\boldsymbol{Y}}, \boldsymbol{U}): \min_{B_{ls}} \sum_{j=1}^{N_e} \| \dot{\boldsymbol{y}}_{k-j} - B_{ls} \boldsymbol{u}_{k-j} \|^2
\end{equation}

\noindent where $\dot{\boldsymbol{Y}} \in \mathbb{R}^{N_e \times 3 n_t}$ and $\boldsymbol{U} \in \mathbb{R}^{N_e \times 3}$ are data matrices containing the last $N_e$ samples of $\dot{\boldsymbol{y}}_k$ and $\boldsymbol{u}_k$, respectively.  

Necessary conditions for a meaningful solution to \eqref{eq: least squares} are given in \cite{zhu_dual-arm_2018} for DLO shape control, including delayed application of \eqref{eq: least squares} until all directions of $\boldsymbol{u}_k$ are present over $N_e$, without which $B_{ls}$ cannot predict dynamics in new directions. As a result, controllers must explore all directions of $\boldsymbol{u}$ to collect sufficient local data, delaying the primary task. 

In section \ref{subsec: sindy dynamics}, rigid body dynamics \eqref{eq: rigid body dynamics}, which assume $\dot{\boldsymbol{y}} = B_{rb}(\boldsymbol{y}) \boldsymbol{u}$, were found to account for a substantial portion of total dynamics. Such a model can predict bulk dynamics in any direction without delay, but does not model more complex local dynamics. To capture the benefits of both approaches and enable instantaneous control without exploration while still learning local dynamics, we propose a residual model: $\hat{f}^{\boldsymbol{y}}(\boldsymbol{y}_k, \boldsymbol{u}_k) = \boldsymbol{y}_k + \Delta t \, \hat{B}(\boldsymbol{y}_k) \boldsymbol{u}_k$, where 
\begin{equation}   \label{eq: ensemble model}
     \hat{B}(\boldsymbol{y}_k) = \alpha_k B_{rb}(\boldsymbol{y}_k) + (1 - \alpha_k) B_{ls}
\end{equation}

\noindent and $\alpha_k \in [0,1]$ indicates the degree to which the rigid body model is favored. To enable immediate control without excitation while gradually incorporating learned local dynamics, let $\alpha_{k+1} = \gamma \alpha_{k}$ with $\gamma \in (0,1)$ and $\alpha_0 = 1$.

To validate \eqref{eq: ensemble model}, for each trajectory in $\mathcal{D}$ we compare $B_{rb}$, $B_{ls}$, and $\hat{B}$. With $N_e = F_c$ (estimate over last 1 second), we predict the next 5 seconds ($5 F_c$ steps); maximum prediction error statistics are shown in Table \ref{table: ensemble model prediction error}, illustrating that the composite model outperforms each constituent model alone. Moreover, the composite model can be deployed without system excitation to immediately apply \textsc{TM-MPC} \eqref{eq: mpc problem}. 

\subsection{Sequence planner} \label{subsec: sequence planner}

While \textsfsmall{TM-MPC} executes a TM, the TM sequence must be planned. We opt for a simple, but effective, heuristic that selects the terminal closest to constraint violation. Among a set of $n_f$ free terminals, the next held terminal is selected with \eqref{eq: terminal selection}.  Finally, the entire \textsfsmall{Install-DLON} procedure including planning and control is given in Algorithm 1. 
\begin{equation} \label{eq: terminal selection}
\begin{aligned}
    \textsfsmall{SELECT}(\boldsymbol{y}; O):  \min_{t \in [0, n_f]} & \min_{o \in O} \quad  (r_{t,o} - \mathbf{d}_{0}(\rho_t, \rho_o)) \\
    \text{s.t.} \quad & \min_{o \in O} (r_{t,o} - \mathbf{d}_{0}(\rho_t, \rho_o)) > 0 \\
                        & \rho_t \in W.
\end{aligned}
\end{equation}
\vspace{-0.5cm}

\begin{algorithm}[b]
\caption{$\textsfsmall{Install-DLON}(\textrm{terms. } \{t\}, \textrm{goals }\boldsymbol{g}, \textrm{obstacles } O)$}
\begin{algorithmic}[1]  
\small
\While{\textsfsmall{True}}  \Comment{Planning loop}
    \State Observe output $\boldsymbol{y}$
    
    \State $t_h \leftarrow \textsfsmall{SELECT}(\boldsymbol{y}; O)$ \eqref{eq: terminal selection}
    
    \If{no solution to \textsfsmall{SELECT} \eqref{eq: terminal selection}}
    
        \State \textbf{break}   \Comment{no feasible TMs remaining}
        
    \EndIf

    \State Grasp terminal $t_h$ with manipulator
 
    \State Initialize $\alpha_0 = 1$, $\dot{\boldsymbol{Y}} \leftarrow \emptyset$, $\boldsymbol{U} \leftarrow \emptyset$
    
    \For{$k \in [0, \overline{N}]$} \Comment{Control loop w/ max. dur. $\overline{N}$}
    
        \State Observe output $\boldsymbol{y}_k$

        \State Estimate $B_{ls} \leftarrow \textsc{LS}(\dot{\boldsymbol{Y}}, \boldsymbol{U})$ \eqref{eq: least squares}

        \State Update model $\hat{B} \leftarrow \alpha_k B_{rb}(\boldsymbol{y}_k) + (1 - \alpha_k) B_{ls}$ \eqref{eq: ensemble model}
                        
        \State Execute $\boldsymbol{u}_k \leftarrow \textsfsmall{TM-MPC}(\boldsymbol{y}_k; \boldsymbol{g}, O)$ \eqref{eq: mpc problem}
        
        \If{$\mathbf{d}_{\beta}(\rho_{h,k}, \rho_{k,g}) < \epsilon_g$}
        
            \State \textbf{break}    \Comment{TM complete}
            
        \EndIf
        
        \State Discount $\alpha_{k+1} \leftarrow \gamma \alpha_k$

        \State Estimate $\dot{\boldsymbol{y}}_k$ and update \textsfsmall{LS} data matrices $ \dot{\boldsymbol{Y}}$ and $\boldsymbol{U}$

    \EndFor
    
    \State Release $t_h$ and return manipulator to home configuration
\EndWhile

\end{algorithmic}
\end{algorithm}

\section{Experiments and Discussion} \label{sec: experiments}

\subsection{Simulated experiments} \label{subsec: simulated experiments}

\begin{figure}
  \centering
  \vspace{0.25cm}
  \includegraphics[width=2.2in]{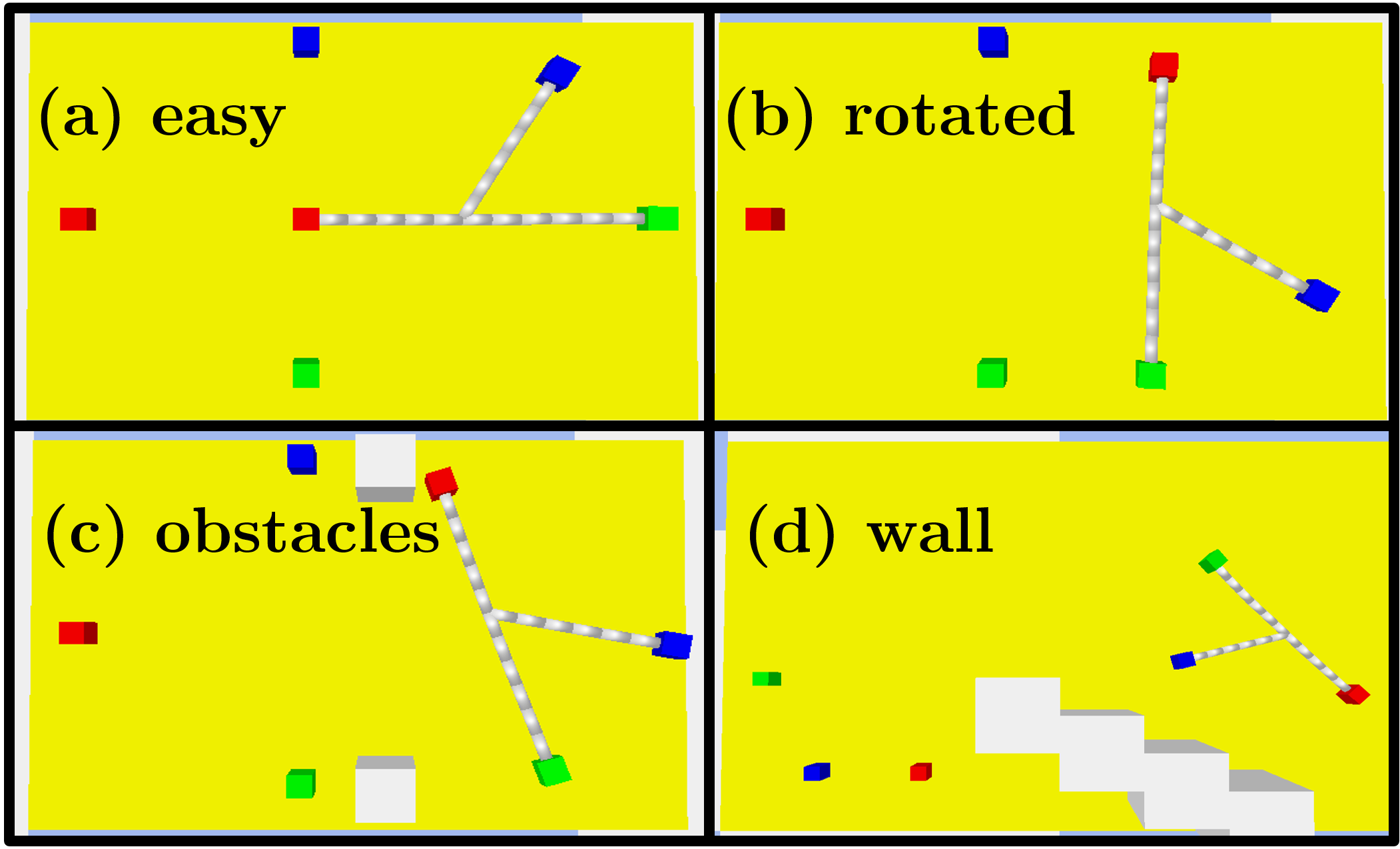}
  \caption{Simulated DLON experiments. Each terminal must be brought to the static receptacle of the same color without leaving the yellow workspace. In (a)-(b), the only obstacles are the receptacles themselves. In (c), the DLON must fit through a narrowing created by obstacles (white). In (d), the DLON must navigate around a blocking wall of obstacles.}
  \label{fig: simulated experiments}
\end{figure}

The DLON model used for dataset collection (section \ref{subsec: dataset collection}) was likewise used for experiments. Recall that our approach does not involve pretraining, so experiments do not benefit from the dataset. Four problems were considered (Fig. \ref{fig: simulated experiments}), in which each terminal of the DLON is to be mated to the receptacle of the same color. \textsfsmall{Install-DLON} was applied to each problem with both the rigid body model \eqref{eq: rigid body dynamics} and the composite model \eqref{eq: ensemble model}, with resulting performance metrics in Table \ref{table: simulated model constraints}: maximum and final constraint values between TMs, where positive values represent constraint violations. 

Problems \textttsmall{easy} and \textttsmall{rotated} involve an obstacle-free workspace. Constraints are respected with both models, though constraints are farther from violation with the composite model. In \textttsmall{obstacles}, obstacles create a narrowing through which the DLON must maneuver. With the rigid model, the DLON becomes stuck and does not finish. With the composite model, a constraint violation is observed, but the problem is solved.  In \textttsmall{wall}, the DLON is initialized behind a wall of obstacles; \textttsmall{wall} is solved with both models, though constraint violations occur that are later recovered.

\begin{table}
    \centering
    \vspace{0.25cm}
    \caption{Simulation problem constraint $c(\boldsymbol{y})$ satisfaction}
    \begin{tabular}{cccc}
        \hline
        Problem & Model Type & Maximum \(c(\boldsymbol{y})\) & Final \(c(\boldsymbol{y})\) \\
        \hline
        \multirow{2}{*}{\textttsmall{easy}} & Rigid Body & $\mathbf{-0.210}$ & $-0.302$ \\
                                   & Composite & $\mathbf{-0.210}$ & $\mathbf{-0.305}$ \\
        \hline
        \multirow{2}{*}{\textttsmall{rotated}} & Rigid Body & $-0.136$ & $-0.264$ \\
                                   & Composite & $\mathbf{-0.143}$ & $\mathbf{-0.290}$ \\
        \hline
        \multirow{2}{*}{\textttsmall{obstacles}} & Rigid Body & \multicolumn{2}{c}{Did not finish} \\
                                   & Composite & \textcolor{red}{$\mathbf{0.031}$} & $\mathbf{-0.015}$ \\
        \hline
        \multirow{2}{*}{\textttsmall{wall}} & Rigid Body & \textcolor{red}{$\mathbf{0.012}$} & $-0.049$ \\
                                   & Composite & \textcolor{red}{$0.017$} & $\mathbf{-0.115}$ \\
        \hline
    \end{tabular}
    \label{table: simulated model constraints}
\vspace{-0.5cm}
\end{table}

\subsection{Physical experiments}  \label{subsec: physical experiments}

\begin{figure}
  \centering
  \includegraphics[width=2.8in]{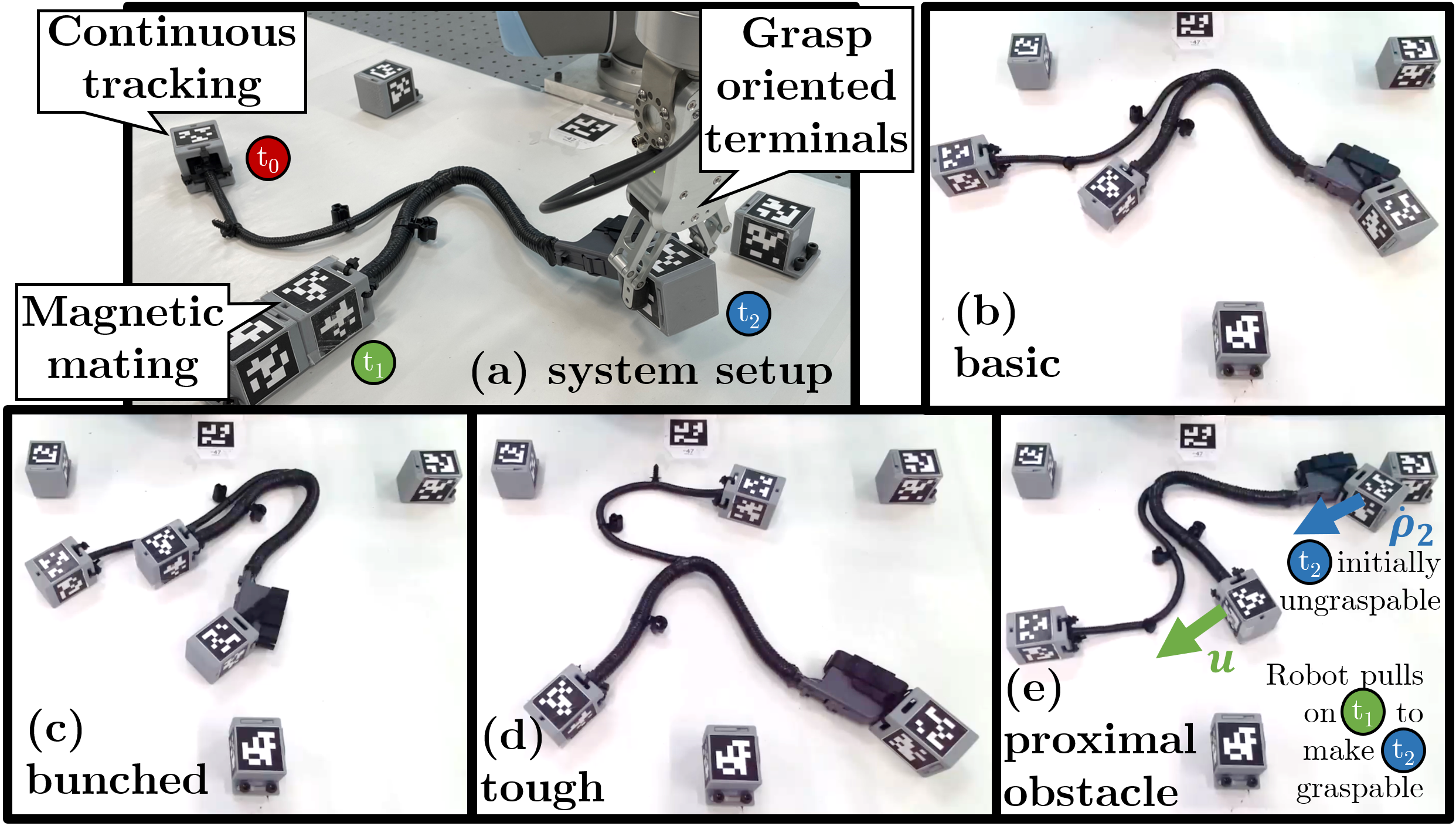}
  \caption{(a) Physical setup with 6-DOF manipulator and modified 3-terminal automotive wire harness. (b-e) Physical experiments with harness configurations in increasing order of challenge. In (b)-(d), terminals are initialized away from obstacles. In (e), terminal $t_2$ is initialized near a receptacle, preventing grasping; the robot grasps  and uses it to pull the violating terminal away from the obstacle before continuing installation. \vspace{-0.6cm}}
  \label{fig: physical experiments}
\end{figure}

To evaluate generalization of \textsfsmall{Install-DLON} and the composite model \eqref{eq: ensemble model}, studies were conducted on a physical system, shown in Fig. \ref{fig: physical experiments}(a). A modification of the setup used in our prior work \cite{Toner2023_RL}, a real automotive wire harness with three terminals, approximately 750 mm in the longest direction, is selected. Blocks of sidelength 45 mm are added at the end of each terminal to ease grasping with a 6-DOF UR5 manipulator with a parallel jaw gripper. AprilTag markers \cite{Wang2016} on terminals and receptacles allow continuous pose tracking with an external Intel Realsense L515 camera. 

The harness was placed in four initial configurations, shown in Fig. \ref{fig: physical experiments}(b-e). In each case, \textsfsmall{Install-DLON} was applied with the composite model, and installation was successful. In the \textttsmall{proximal-obstacle} problem, terminal $t_2$ is placed very close to a receptacle, preventing grasping. The robot selects $t_1$ to hold, which it uses to pull $t_2$ out of constraint violation before moving on to the insertion of $t_1$, thus ensuring feasibility of a later $t_2$ insertion.

\subsection{Discussion of results} \label{subsec: results discussion}

The experimental results indicate the suitability of our planning and control approach for solving the DLON installation problem, as defined in section \ref{sec: problem statement}. MPC naturally allows incorporation of the constraints required by TMs, and the main challenge becomes selection of a sufficiently accurate dynamics model for the MPC. Simulations indicate that the adaptive composite model outperforms the rigid body model in terms of predicting constraint violation, where the latter fails entirely in one problem. The composite model, which was originally developed with insights from simulation data, generalizes to a new, physical DLON, enabling prediction of free terminal dynamics suitable for effective MPC. 

Our proposed approach is capable of manipulating a DLON to endpoint goals without monitoring its internal state, and reveals that the complex, data-driven global models \cite{zhu_dual-arm_2018} developed for DLO state control may not be necessary for endpoint control. Compared to DLO endpoint control policies learned offline from real data \cite{lim2022_casting}, our composite model predicts free terminal trajectories sufficiently accurately for control purposes without offline learning.

\section{Conclusions and Future Work} \label{sec: conclusions}

In this work, planar manipulation of DLO networks (DLONs), collections of connected stiff DLOs (StDLOs), is studied. Focus is directed toward DLONs with measured terminal poses (output) but unmeasured configurations (state). The existence of output dynamics without explicit state dependence is demonstrated in simulation. The output dynamics are approximated, revealing significant rigid body motion. This structural insight leads to a adaptive composite model, consisting of a simple rigid body model and a continuously estimated local linear model, which is used by an MPC for constrained, single-robot DLON terminal manipulation (TM). A simple TM sequence planner based on constraint proximity is proposed. The efficacy of our framework is demonstrated in simulated and physical experiments. 

This study demonstrates the feasibility of output-based control on a class of planar DLON problems, but practical problems are more complex. Assumption (A1) prohibits state constraints, preventing the framework from being deployed in environments with obstacles likely to collide with the DLOs during a TM. Moreover, although predefined grasp points may be necessary to prevent DLON damage, restricting grasps to terminals is limiting. Future work towards explicit state estimation would relax (A1) and expand candidate grasp poses, enabling more complex manipulation sequences. 

Although our method is not constrained in principle by DLON complexity, it was only validated on 3-terminal DLONs with similar characteristics. Future work is needed to validate our models and understand the limitations of our planning and control approach. In particular, understanding the range of physical properties that qualify as an StDLO, such that the output map is approximately invertible, is critical when only sparse terminal pose measurements are available. Additionally, extending the approach to non-planar DLONs will be necessary for some applications, such as harness installation within an automotive door assembly.

As harness installation is a repetitive task, future work will investigate how data from repeated DLON installations could be leveraged to build more complex models from real data \textit{in situ} without deliberate excitation, as in \cite{wang_offline-online_2022, lim2022_casting}, to better inform both control-oriented models and high-level plans. Such an approach could use our local, DLON-agnostic composite model to enable initial installations without \textit{a priori} data, while allowing performance gains over time from more accurate global models. 




\section*{Acknowledgement}
\small{
\noindent This material is based upon work supported by the National Science Foundation Graduate Research Fellowship under Grant No. DGE-1841052. Any opinion, findings, and conclusions or recommendations expressed in this material are those of the authors(s) and do not necessarily reflect the views of the National Science Foundation. 
}


\bibliographystyle{ieeetr}
\bibliography{references}

\begin{thebibliography}{10}

\bibitem{saha_manipulation_2007}
M.~Saha and P.~Isto, ``Manipulation {Planning} for {Deformable} {Linear} {Objects},'' {\em IEEE Transactions on Robotics}, vol.~23, pp.~1141--1150, Dec. 2007.

\bibitem{moll_path_2006}
M.~Moll and L.~Kavraki, ``Path planning for deformable linear objects,'' {\em IEEE Transactions on Robotics}, vol.~22, pp.~625--636, Aug. 2006.

\bibitem{Jiang2010}
X.~Jiang, K.-m. Koo, K.~Kikuchi, A.~Konmo, and M.~Uchiyama, ``{Robotized Assembly of a Wire Harness in Car Production Line},'' in {\em International Conference on Intelligent Robots and Systems}, 2010.

\bibitem{lv_dynamic_2022}
N.~Lv, J.~Liu, and Y.~Jia, ``Dynamic {Modeling} and {Control} of {Deformable} {Linear} {Objects} for {Single}-{Arm} and {Dual}-{Arm} {Robot} {Manipulations},'' {\em IEEE Transactions on Robotics}, vol.~38, pp.~2341--2353, Aug. 2022.

\bibitem{wang_offline-online_2022}
C.~Wang, Y.~Zhang, X.~Zhang, Z.~Wu, X.~Zhu, S.~Jin, T.~Tang, and M.~Tomizuka, ``Offline-{Online} {Learning} of {Deformation} {Model} for {Cable} {Manipulation} {With} {Graph} {Neural} {Networks},'' {\em IEEE Robotics and Automation Letters}, vol.~7, pp.~5544--5551, Apr. 2022.

\bibitem{wu2020learning}
Y.~Wu, W.~Yan, T.~Kurutach, L.~Pinto, and P.~Abbeel, ``Learning to manipulate deformable objects without demonstrations,'' in {\em 16th Robotics: Science and Systems, RSS 2020}, MIT Press Journals, 2020.

\bibitem{yan2021learning}
W.~Yan, A.~Vangipuram, P.~Abbeel, and L.~Pinto, ``Learning predictive representations for deformable objects using contrastive estimation,'' in {\em Conference on Robot Learning}, pp.~564--574, PMLR, 2021.

\bibitem{han_model-based_2017}
H.~Han, G.~Paul, and T.~Matsubara, ``Model-based reinforcement learning approach for deformable linear object manipulation,'' in {\em 2017 13th {IEEE} {Conference} on {Automation} {Science} and {Engineering} ({CASE})}, (Xi'an), pp.~750--755, IEEE, Aug. 2017.

\bibitem{alvarez_interactive_2016}
N.~Alvarez and K.~Yamazaki, ``An interactive simulator for deformable linear objects manipulation planning,'' in {\em 2016 {IEEE} {International} {Conference} on {Simulation}, {Modeling}, and {Programming} for {Autonomous} {Robots} ({SIMPAR})}, (San Francisco, CA, USA), pp.~259--267, IEEE, Dec. 2016.

\bibitem{fresnillo_approach_2022}
P.~M. Fresnillo, S.~Vasudevan, and W.~M. Mohammed, ``An approach for the bimanual manipulation of a deformable linear object using a dual-arm industrial robot: cable routing use case,'' in {\em 2022 {IEEE} 5th {International} {Conference} on {Industrial} {Cyber}-{Physical} {Systems} ({ICPS})}, (Coventry, United Kingdom), pp.~1--8, IEEE, May 2022.

\bibitem{zhu_robotic_2020}
J.~Zhu, B.~Navarro, R.~Passama, P.~Fraisse, A.~Crosnier, and A.~Cherubini, ``Robotic {Manipulation} {Planning} for {Shaping} {Deformable} {Linear} {Objects} {With} {Environmental} {Contacts},'' {\em IEEE Robotics and Automation Letters}, vol.~5, pp.~16--23, Jan. 2020.

\bibitem{yu2021adaptive}
M.~Yu, H.~Zhong, F.~Zhong, and X.~Li, ``Adaptive control for robotic manipulation of deformable linear objects with offline and online learning of unknown models,'' {\em arXiv e-prints}, pp.~arXiv--2107, 2021.

\bibitem{lim2022_casting}
V.~Lim, H.~Huang, L.~Y. Chen, J.~Wang, J.~Ichnowski, D.~Seita, M.~Laskey, and K.~Goldberg, ``Real2sim2real: Self-supervised learning of physical single-step dynamic actions for planar robot casting,'' in {\em 2022 International Conference on Robotics and Automation (ICRA)}, pp.~8282--8289, 2022.

\bibitem{zhou_practical_2020}
H.~Zhou, S.~Li, Q.~Lu, and J.~Qian, ``A {Practical} {Solution} to {Deformable} {Linear} {Object} {Manipulation}: {A} {Case} {Study} on {Cable} {Harness} {Connection},'' in {\em 2020 5th {International} {Conference} on {Advanced} {Robotics} and {Mechatronics} ({ICARM})}, (Shenzhen, China), pp.~329--333, IEEE, Dec. 2020.

\bibitem{koo2008}
K.~M. Koo, X.~Jiang, K.~Kikuchi, A.~Konno, and M.~Uchiyama, ``{Development of a robot car wiring system},'' in {\em IEEE/ASME International Conference on Advanced Intelligent Mechatronics, AIM}, pp.~862--867, 2008.

\bibitem{zhu_dual-arm_2018}
J.~Zhu, B.~Navarro, P.~Fraisse, A.~Crosnier, and A.~Cherubini, ``Dual-arm robotic manipulation of flexible cables,'' in {\em 2018 {IEEE}/{RSJ} International Conference on Intelligent Robots and Systems ({IROS})}, pp.~479--484, {IEEE}.

\bibitem{Nguyen2021}
T.~P. Nguyen and J.~Yoon, ``{A novel vision-based method for 3D profile extraction of wire harness in robotized assembly process},'' {\em Journal of Manufacturing Systems}, vol.~61, pp.~365--374, oct 2021.

\bibitem{Toner2023_RL}
T.~Toner, M.~Saez, D.~M. Tilbury, and K.~Barton, ``Opportunities and challenges in applying reinforcement learning to robotic manipulation: an industrial case study,'' {\em Manufacturing Letters}, 2023.

\bibitem{Saez2020}
M.~Saez and P.~Spicer, ``{Robot-to-robot collaboration for fixtureless assembly: Challenges and opportunities in the automotive industry},'' in {\em ASME 2020 15th International Manufacturing Science and Engineering Conference, MSEC 2020}, vol.~2, pp.~8--10, 2020.

\bibitem{Zhao2022}
T.~Z. Zhao, J.~Luo, O.~Sushkov, R.~Pevceviciute, N.~Heess, J.~Scholz, S.~Schaal, and S.~Levine, ``{Offline Meta-Reinforcement Learning for Industrial Insertion},'' in {\em International Conference on Robotics and Automation (ICRA)}, 2022.

\bibitem{coumans2019pybullet}
E.~Coumans and Y.~Bai, ``Pybullet, a python module for physics simulation for games, robotics and machine learning.'' \url{http://pybullet.org}, 2016--2019.

\bibitem{Akiba2019}
T.~Akiba, S.~Sano, T.~Yanase, T.~Ohta, and M.~Koyama, ``{Optuna: A Next-generation Hyperparameter Optimization Framework},'' in {\em Proceedings of the ACM SIGKDD International Conference on Knowledge Discovery and Data Mining}, pp.~2623--2631, 2019.

\bibitem{brunton_discovering_2016_sindy}
S.~L. Brunton, J.~L. Proctor, and J.~N. Kutz, ``Discovering governing equations from data by sparse identification of nonlinear dynamical systems,'' {\em Proceedings of the National Academy of Sciences}, vol.~113, pp.~3932--3937, Apr. 2016.

\bibitem{brunton_sparse_2016_sindyc}
S.~L. Brunton, J.~L. Proctor, and J.~N. Kutz, ``Sparse {Identification} of {Nonlinear} {Dynamics} with {Control} ({SINDYc}),'' {\em IFAC-PapersOnLine}, vol.~49, no.~18, pp.~710--715, 2016.

\bibitem{de2020pysindy}
B.~de~Silva, K.~Champion, M.~Quade, J.-C. Loiseau, J.~Kutz, and S.~Brunton, ``Pysindy: A python package for the sparse identification of nonlinear dynamical systems from data,'' {\em The Journal of Open Source Software}, vol.~5, no.~49, p.~2104, 2020.

\bibitem{Wang2016}
J.~Wang and E.~Olson, ``{AprilTag 2: Efficient and robust fiducial detection},'' in {\em 2016 IEEE/RSJ International Conference on Intelligent Robots and Systems (IROS)}, pp.~4193--4198, IEEE, Oct 2016.

\end{thebibliography}

\end{document}